School of Computer Science and Information TechnologyUniversity of Nottingham
Jubilee Campus
NOTTINGHAM NG8 1BB, UK

# Memory Implementations - Current Alternatives

*William Wilson and Dr Uwe Aickelin*

# Memory implementation – Current Alternatives.

William Wilson
Supervised by Dr U. Aickelin
ASAP Group, University of Nottingham, UK
Email: wow@cs.nott.ac.uk

June 9, 2005

## 1. Introduction

Memory can be defined as the ability to retain and recall information in a diverse range of forms. It is a vital component of the way in which we as human beings operate on a day to day basis. Given a particular situation, decisions are made and actions undertaken in response to that situation based on our memory of related prior events and experiences. By utilising our memory we can anticipate the outcome of our chosen actions to avoid unexpected or unwanted events. In addition, as we subtly alter our actions and recognise altered outcomes we learn and create new memories, enabling us to improve the efficiency of our actions over time. However, as this process occurs so naturally in the subconscious its importance is often overlooked.

The implementation of memory in a problem solving domain offers an attractive proposition but it is often underutilised. In most applications memory consists of simply recording all information that is generated from a process. One could argue that computers have developed to such a degree that storage issues are no longer a concern, therefore why not store every bit of data generated? I believe however, that in doing so one is missing out not only on a unique and efficient storage mechanism but also an important data retrieval process and a key driver in the functionality of the system.

Without incorporating a true representation of memory I believe the system is created with artificial constraints that limit its potential. A prime example of this can be seen in Artificial Immune Systems (AIS). Thus far memory implementations in AIS's have been very simplistic; however memory in the natural immune system is incredibly complex and is recognised as a key player in driving the immune response to counter virus re-infections. AIS's have so far failed to "stand out from the crowd" in terms of their performance and functionality and I believe a key reason for this is the simplistic implementation of memory. How can an artificial system offer the same attractive properties as its natural counterpart if something as key as memory has been relegated to such a simplified role.

I believe therefore that memory is key in any system implementation. The following review has been performed to firstly investigate the desirable properties of such a memory mechanism. Discussion then moves onto an analysis of the current methods of incorporating memory into a variety of problem solving domains. Each of these will be critically evaluated to identify the benefits and issues that such an implementation would offer when taken and considered in terms of our problem domain. The problem being investigated by our team is the implementation of an intrusion detection system (IDS) operating on a network of computers. The objective of the IDS will be to detect and report any worm based intrusions that occur on the network. A worm can be identified by its signature. The purpose of memory in this system is to ensure that knowledge gained on a previous signature experience can be recalled and used to ensure a faster response when that same or similar signature is re-experienced; we get a faster more effective secondary response which leads to a quicker recognition of a known worm virus or a mutation of that virus.

## 1.1 Desirable properties under consideration for a memory mechanism

The properties of memory that are desirable for such an IDS system can be prioritised as follows:

- **Generalisation** – Can the system take the specific properties of a signature from a computer worm virus and be able to abstract them into a more generalised entity for storage as memory? The system is then able to use this generalised memory as a starting point and search around the local shape space of that initial point to recognise mutations of a specific signature. In this way the system does not have to remember every single signature variation but a generalised abstraction of them. The system has the ability to associate a novel signature presented with one of its existing memory items.
- **Adaptation** – When attempting to identify and recognise a novel signature, is the system able to take its own solutions to recognising that signature, and if they are not a good fit dynamically adapt them in an attempt to improve their affinity; or is this adaptation process driven from an external process outside of the system? Is the system able to start off with an empty memory set which can develop dynamically over time based on the information learnt rather than having to be informed of what to remember when the system is initially created?
- **Persistence** – Can the system forget information that it has previously memorised? Without repeated stimulation some systems may tend to forget information relating to older less frequently occurring viruses. This could have serious repercussions for an IDS.
- **Scalability & Efficiency** – Assesses the ease with which the system can be scaled in size to incorporate larger population sizes. Are signature items stored in an efficient way to ensure that the system can be scaled up with minimal impact on performance?
- **Memory extraction** – Is it feasible to extract the entity that encapsulates the principle of memory in the system and be able to analyse and utilise it?
- **Unbiased** – Does the system treat all memories accumulated in an unbiased way or does it tend to favour remembrance of the most common viruses, or those that have the closest match to its population set?
- **Accuracy** – Is the system able to accurately classify novel viruses based on its existing memory pool to minimise the occurrence of incorrect solutions?
- **Ease of implementation** – Is the system simple and easy to develop, implement and maintain?

A number of current memory implementations will now be discussed to assess whether they incorporate these desirable properties.

## 2. Memory Implementations.

### 2.1 Complexity Theory and Emergence.

A complex system consists of a large number of individual entities that taken in isolation have very simplistic function and ability. They interact at a very basic level in their local environment, having no real concept of the system as a whole in which they are participating. However, due to the interactions between these entities at the local level, certain behaviour patterns emerge as a property of the system as a whole [1]. A prime example of one such emergent property is memory. At the individual level entities may have little capability for memory maintenance but through their collective behaviour the population as a whole develops a propensity to remember information. This arises due to the feedback loops generated during the interactions of individual population members.

A simple example of a complex system is an ant colony [2,3]. Each ant follows a very simple behaviour pattern in its local environment; it has little if any awareness of the size, objectives or behaviour of the colony. Yet through the interactions of individual ants a collective behaviour emerges for the colony as a whole. An individual ant will live for a matter of days however the ant colony is able to remember where sources of food exist over the course of years. This memory is not instilled in individual ants within the colony but is created and sustained via the constant interactions between the ants. These interactions are possible because of pheromones [3].

While walking between food sources and the nest ants deposit a substance called *pheromone*, forming a pheromone trail. Ants can smell pheromone and, when choosing their way, they tend to choose paths marked by strong pheromone concentrations and in this way find the location of the food sources found by their nestmates [2]. When a colony is created, all the ants will commence hunting for food by following random paths and depositing pheromones on the path they choose. The ant that finds food first will then attempt to return to the colony; the ant will automatically follow his original path back as that is the only one to contain any pheromone. This increases the level of pheromone on this path further and so will attract other ants, who in turn deposit more pheromone on this path, thereby creating a reinforcement feedback effect.

Even though the individual ants may die over time, the pheromone path created via the ants' interaction remains and memory of the food source emerges as a global property of the colony. In this analogy memory is conceptualised as the actual pheromone trail as this trail represents the information learnt by the system through its past operations. The knowledge encapsulated by the pheromone trail evolved dynamically and is used to direct the activity of the colony over time.

Similarities can be seen between ants in a colony and cells in the immune system. It is difficult to comprehend that individual immune cells adapt to retain the memory of a specific pathogenic source, as the lifespan of most cells is relatively short. The risk of losing that one cell, and therefore the loss of knowledge of the infection is too high. In addition there are not enough cells in the human body to store all possible pathogenic derivatives. It would appear more logical that memory is not attributed to individual cells but arises as an emergent property from the interactions and communication of those immune cells.

If we use the concept of 'Shape Space' [4] then a point in 'N' dimensional shape space could represent 'N' characteristics associated with an antigen. The concept of pheromone trails could be used to direct the hyper mutation process in that shape space. Immune (B) cells located in one area of shape space would mutate to become a better fit to the antigen and in so doing would move across that shape space. Using this pheromone analogy, cells undergoing successful mutations would encourage other cells to follow the same mutation path. This would lead to a quick global convergence on the pathogen source.

Considering our IDS application this approach would imply the system could quickly adapt to identify and retain knowledge of strong virus types in a dynamic autonomous manner. Memory would not need to be encoded in individual components but would arise naturally as a property of the workings of the system. Memory here would be encoded as the most successful mutation path in shape space.

However there are a number of problems associated with this approach.

- **Memory abstraction** – memory is a property of the working of the system not an attribute of the components in the system, therefore it would not be easy to extract and utilize that memory. Using the ant colony analogy there is no explicit record of the pheromone trail to the food source, it arises from the behaviour of the ants. Considering our application it would therefore be difficult to explicitly identify the actual mutation path from a generic to the actual virus source.
- **Memory bias** – Memory would develop of only the strong food / virus sources as ants are encouraged away from other potential sources by the more dominant pheromone trails. Memory of smaller food sources would diminish over time or be missed in the first place. Considering our application this means that less significant / frequent viruses could be easily missed by the system.
- **Pheromone evaporation** – Over time the pheromone trail evaporates for two reasons: to prevent premature convergence to a local optima, and to allow ant colonies to slowly forget their past history so that they can direct their search in new directions. This means that over time the system can forget; this would be undesirable in an IDS, as memories of less significant viruses would inevitably disappear.
- **Inability to adapt to new viruses** – The ant approach would be less likely to detect newly arising viruses because they would be focussed on reinforcing the paths to the most significant virus sources.
- **Difficulty in modelling** – As you don't explicitly model memory, it emerges naturally from the system how do you ensure it occurs in the system in the way you want it to? Control and regulation of memory is taken out of the users hands, creating uncertainty.

## 2.2. Neural Networks – Hopfield network.

A Hopfield neural network consists of a large number of fully interconnected nodes which act as processing units. Each interconnection has an associated weight and these are randomly generated when the network is established. Each node has a state, and this state can change based on the total value of the connections that feed into that node. The vector of the states of each node establishes a state for the overall network.

Training data is fed to the network and each node then undergoes an update procedure one at a time. If the output of the network does not match the input data then the update procedure adjusts the relevant weights in an attempt to create a match. The process is repeated for every node and then repeated for the whole network numerous times for each item in the training set. With a suitable update procedure the network will converge and stabilise. Once it has stabilised it can be used to perform such tasks as optimisation and associative memory [5].

Unlike the human brain the Hopfield network has to be told what to remember, in this sense the training data can represent the items you wish the network to remember. After presentation of the memory set the update procedure adjusts the weights of the network to lead to a convergence of the network on the memory set. Memory in the system is encapsulated in the weights allocated to the interconnections and these can be extracted, however they are meaningless without the network used to create them.

Memory representation can be visualised in terms minimum points in a 2d landscape [6].

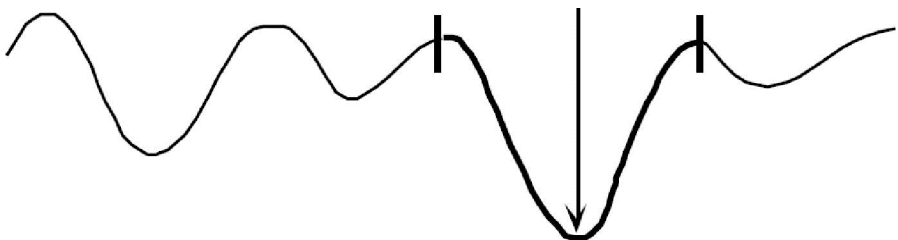

**Figure 1. Memory representation in a 2D landscape [6].**

Minimum values for the state of the network correspond to areas where the network converges on one of the memory set items. Each minimum point on the landscape has an associated basin of attraction [6]. When presenting a novel data item to the network the network starts off in a location on this landscape. If the network starts off on the slope of one of the basins of attraction then the update procedure moves the state of the network until it is at a minimum, this then corresponds to one of the memories of the network associated with that basin. The basin of attraction is therefore a generalisation of the originally imprinted memory item. Through this generalisation the network can associate the novel item presented with one of the memory classifications that have been imprinted.

This ability to generalise is advantageous for intrusion detection systems. The system does not have to remember all derivatives of a particular virus, just one generic one. If altered versions of the virus are then presented, the network will be able to associate these mutated viruses to their original sources. The solution relevant to the generic virus can then be applied to this new derivative. Memory is therefore more efficient in terms of the resources needed to maintain it.

However a number of issues arise with this approach:

- **Generalisation and scalability** – If the basins of attraction are too wide then the network would associate some novel viruses as derivates of a virus to which they don't belong i.e. we get over-generalisation. This association could mean that inappropriate corrective behaviour is applied to the presented virus. This problem arises because the basins of attraction are created from the memory set that is initially presented to the network. This memory set is unlikely to cover all possible future alternatives as new viruses arise all the time that may not relate to the ones in the original memory set. Alternatively if there are too many basins of attraction we get excessive convergence and too much detail. To avoid inappropriate classifications the network would have to be regularly retrained to generate additional basins of attraction to reflect the diversifying virus range. This process creates a problem of scalability.
- **Limited capacity** – As we imprint more patterns on the network the number of basins of attraction will increase. There will be a limit on the number that can be added before the basins of attraction begin to destructively interfere with each other [6], making association difficult. It was found that the total number of memories = $0.15N$ [5] where $N$ = the total number of processing units. Given this, an increase in memory capacity requires a more than proportional increase in the population in the system; this appears highly inefficient and leads to large complex systems that are difficult and costly to train.
- **Incorrect association** – Experiments using Hopfield networks have shown that on occasion the memory evoked as being associated to the input pattern of the item presented is not necessarily the memory pattern that is most similar to that input pattern [5]. The journey down the basin of attraction did not happen in the way it should have done. However this level of inaccuracy will be intrinsic in all approaches.
- **Inconsistent emphasis on different memory patterns** – all memories are not remembered with equal emphasis as basins of attraction may differ in size and shape.
- **Predefined memory pool** – Upon creation the user would have to identify a specific memory set for the system to remember, the system does not have the capability to identify and memorise new unique viruses unless the training process was repeated using these novel items, which could take considerable time.

## 2.3. Artificial immune systems.

Due to the sheer volume and complexity of the mechanisms and interactions that exist within the immune system, knowledge of the exact operation of immune memory is limited. As a result implementations of immune memory have varied considerably based on the information gleaned from the immunological community. A selection of those implementations is described below.

a) **Long lived memory cells**.

In this theory memory of an antigen encounter is embedded in a single immune cell called a memory cell that lives for an indefinite time and can be used as a source of reference when subsequent antigen are presented to the system. During the initial presentation of an antigen, highly attuned immune cells that bind to the antigen will clone, and some of those clones differentiate into long lived memory cells that carry forward the knowledge of that experience in case of a re-infection.

The concept of long lived cells that retain knowledge of an antigen encounter has been implemented in many AIS models including CLONALG [11], aiNET [15], MLAIS [12], AIRS[14], and Hunt's model [16]. In these approaches it is common to have a separate distinct memory pool which is initially empty. As antigens are presented to the system the highest affinity immune cell to the antigen is identified and a complimentary or similarity match is made between this cell and the antigen. In most cases, especially those based on B cell behaviour, the immune cell is given the chance to mutate to improve its affinity with the antigen. Over time the match between the immune cell and antigen becomes such that the two bind and the antigen is removed. The time is takes for this mutation and binding process creates the time lag associated with a primary response.

After binding the immune cell clones itself to increase the population of affinity attuned immune cells and one of those clones is selected as a memory candidate. The memory candidates are compared to those in the memory pool and if the pool is not full the memory cell is entered into the memory pool. If the pool is full then the memory candidate is compared to the memory cell that exhibits the highest affinity to the current antigen. The method of comparison varies according to the implementation but generally if the memory cell has a higher affinity than the memory candidate then it stays in the memory pool, otherwise the memory candidate will replace it. Over time the memory pool evolves to counteract the most frequent and recent antigen. To reflect reality the size of the memory pool is limited, so it is not possible to maintain a copy of the best fitting immune cell from every antigen encounter.

Upon secondary presentation of an antigen the comparison is made again against all immune cells including the memory cells. As these are already attuned to the antigen the bind will be significantly quicker, as there is no lag required for mutation. In addition as the degree of clonal expansion is directly related to the affinity of the bind there will be a faster, larger clonal expansion indicative of a secondary response.

The concept of a long lasting memory pool consisting of cells that retain knowledge of prior antigen experiences has many advantages; it is simple to implement, it is able to generalise because of the threshold associated with binding, and it can adapt over time to new antigen threats because of the diverse population of immune cells and their ability to mutate to match new threats. In addition because the memory pool is separately maintained from the rest of the population it can be easily extracted for analysis at any time. Because of the limited size of the memory pool, and its ability to generalise, it is also efficient at storing information.

However there are a number of issues with such a representation:

- **Persistence and bias**: This represents the biggest issue to the system as the memory pool is biased towards remembering only the most recent or frequent antigens presented to it. Due to the competition among memory cells and memory candidates, older memory cells associated with antigen presented in the past will likely be replaced by newer memory candidates that detect the most current antigen. In this way the memory pool evolves but in doing so it forgets.
- **Scalability**: The memory pool is always going to be limited in size in proportion to the overall population level and as we increase the scale of the system the ability of that pool to cover the diversifying antigen population will become more limited.
- **Implementation**: The ability of this system to generalise depends on the valuation of the mutation rate and bind threshold parameters, it is therefore critical that these be determined appropriately. If the recognition regions associated with antigen binding are too large then the system over generalises and the results are meaningless.
- **Biological feasibility**: In trying to build an AIS it is important to simulate reality to develop behaviour that provides the advantages of such a natural system. Current immunologists however stress that long lived immune cells are unlikely to exist [10].

b) **Idiotypic Networks**.

This theory recognises that individual immune cells do not have an indefinite lifespan, and therefore something must continuously stimulate successful immune cells to encourage them to divide so they can pass on knowledge of their antigen experience to their progeny. Knowledge gained during an immune response is therefore carried forward through the generations of immune cells and is not held in one cell for an indefinite period.

Different implementations use different method to induce this homeostatic turnover. Some use a residue of the original antigen to trigger a constant stimulation to the memory pool, however this has been dismissed as a biological reality. In addition it also implies that a record of each antigen source must be maintained, which makes a system to remember such information pointless. An interesting alternative is that of the idiotypic network, created by computer scientists based on their interpretation of how the immune system could theoretically work.

The principle behind the idiotypic approach is that it assumes immune cells will interact with each other as well as interacting with antigen, and that this interaction will influence the behaviour of the system and could act to stimulate homeostatic turnover. For example it is feasible that an immune cell 'A' may reflect the internal representation of an antigen source that another immune cell B would react to. 'B' reacts to 'A' and is stimulated to proliferate. This proliferation of B suppresses the expansion of cell type 'A'. However another cell type 'C' may react to 'B's expansion, leading to further suppression and stimulation. These loops of stimulation and suppression create a dynamic mechanism that ensures the memory of the original encounter is propagated [17] by a stimulation pulse that bounces back and forth amongst the interacting cells

The simple idiotypic network comprised a linear chain, with a beginning and an end to the chain, enabling only a restricted interaction between cells, however these limitations were somewhat expanded by linking the ends of the linear chains to create cyclic networks where all the cells interacted in a continuous loop [17]. This provided a more complex, yet still simplified, representation of what was considered possible in the real system. Further developments were made by creating Cayley tree like networks [17], expanding the number of connections between the artificial cells whilst maintaining the loops of stimulation and suppression. These networks were successful, were seen to dynamically evolve, and were able to maintain memory.

As idiotypic networks are so similar to neural networks they benefit from the same advantages of such systems, including the ability to generalise and adapt, however they suffer all the same disadvantages, and a few unique issues, due to the nature of their construction as defined below:

- **Instability**: It was noted [18] that if the cyclic network consisted of an odd number of nodes then frustration was likely to develop. Frustration describes the situation when the loops of stimulation and suppression led to instability in the network. In order for the network to resolve the instability nodes had to be dropped from the network, creating the potential for memory cells to be lost in the process.
- **Scalability**: Due to the complex nature of the interactions possible, any scaling up in the magnitude of the system would have serious repercussions for its performance; this represents one of the most significant factors for such network based solutions.
- **Biological implausibility**: Both computer scientists and immunologists recognise that idiotypic networks have no basis in biological truth. Although this should not prevent us looking at it as a potential solution, further research has shown that the use of network based approaches are ultimately limited in an AIS domain.

**c) Population emergent memory.**

Other systems have recognised the benefits of maintaining memory capability within the dynamics of a cell population. These systems work in a similar manner to the long lived memory cell theory but they do not extract the high affinity cells to form part of an elite set. Instead these theories let the cell populations mutate, bind, clonally expand, differentiate, and die as a part of a naturally evolving system that responds to antigen presentations. Memory therefore evolves dynamically as part of the population behaviour. All cells follow the same rules set and memory properties evolve over time as successful cells clonally expand to fill a larger proportion of the population set. The proportion of cells that are attuned to an experienced antigen source therefore increases after exposure. Given homeostatic conditions, each cell has a similar probability of dying, therefore over time knowledge of the antigen source is likely to be carried forward as there are a larger number of cells associated with that antigen. Any re-exposure is likely to be met by this attuned subpopulation leading to a faster, more effective immune response. We can see therefore that memory has evolved as a characteristic of the population as a whole and not as a characteristic within individual cells [19,20].

This process creates a system that is highly flexible, able to adapt to a diverse antigen range. It is a truly dynamic system where the population evolves to counter any threat made to it. This would prove advantageous in an IDS as the recognition of viral signatures is not static and cannot be predefined. Signatures change at an extremely rapid rate in unpredictable ways to seek new exploits in the system. To cope with this rapid development the IDS has to be extremely flexible and as a result so does the memory mechanism, and this approach satisfies this requirement.

One problem these approaches have however is how to maintain the knowledge of antigen experience over time. Due to its evolving nature the population of attuned cells will divert away from earlier antigen sources towards those that are more recent / frequent / severe, and so there is a risk that the system will forget earlier antigen experiences unless they are continually reintroduced. Solutions employed to maintain this memory vary according to each implementation but consist of some of the following:
- antigen persistence
- natural turnover of the memory pool
- competition for resources re density dependent death rate
- variations in death rates.
- signal dampening
- telomeres and telomerase.

Although complex to implement this evolutionary form of memory is an extremely attractive proposition for an IDS due to its ability to generalise in an efficient and effective manner that ensures the existence of a core generic memory pool that can react to an almost infinite variety of possible virus derivatives, thus leading to a more efficient use of resources. The system is also dynamic unlike neural networks. The system will automatically attempt to remember new novel items presented without the need to retrain, it evolves over time based on the information presented and can evolve from an empty memory set.

However there are a number of issues that remain with this form of memory and its use in an IDS:

- **Efficiency**: AIS's in general require a vast pool of resources in order for the system to run effectively. There needs to be a huge immune cell population that has the ability to mutate to recognise novel antigen. However after antigen presentation only a small percent of the population are used in the immune and memory response, creating huge inefficiencies in system operation.
- **Implementation**: Due to its dynamic nature the system is complex to implement as you are trying to model complex population dynamics and interactions. Mechanisms will also have to be chosen to introduce a homeostatic mechanism – these will be subjective and have a great influence on the systems behaviour.
- **Persistence**: Without the incorporation of a homeostatic mechanism memory of an antigen experience may be forgotten over time, which would have a significant impact on the system.
- **Memory extraction**: As memory is an emergent property of the immune cell population it may be difficult to extract memory associated with explicit antigen presentations. Memory of that interaction would have been carried forward across a number of generations of cells which may have subsequently mutated to counteract other antigen. As a result this system has become more dynamic but also more difficult to extract explicit information from.
- **Scalability**: Because of the complex nature of the workings of the system, issues will arise when the model is scaled up to accommodate larger data sets.

## 2.4. Hash tables.

Hash tables are ideal data structures for use when you have large amounts of data and you need to locate one of those data items very quickly. Memory in this system is created by storing each item that is presented in a unique location. The data structure used to house all the data items is much like an array. A hash function is then used to convert the data item itself into a reference that corresponds to the address of the element in the array where that data item is to be stored. You can then access the data item directly by passing it (the data item) through the hash function to obtain the location in the array where it is stored, without having to search the array element by element.

This would be ideal in an IDS as data presented to the system can be quickly referenced to the hash table to identify whether they are a virus or not. Speed is an essential requirement of the IDS so a hash table is ideal in this sense.

In addition, in the other models discussed there are excessive resources required to maintain memory of a single virus. In neural networks you would need to increase the number of nodes by 10 [5] to incorporate an additional memory pattern. With an AIS there is huge redundancy as most entities in the model do not participate in the clonal selection and memory process. Here one object is required to maintain the memory of one item and so at first glance it proves very efficient.

However there are a number of disadvantages to this approach:

- **Inability to generalise** – as new viruses are found they will have to be added to the hash table. If each virus is slightly different, passing them through the hash function will generate different addresses for storage and therefore identify them as separate individual items. In reality they may be similar viruses that are derivations of the same virus source. Instead of being grouped for the purposes of a relevant solution the system cannot generalise, creating a less efficient storage system as relationships between entries are not be easily observable.
- **Complexity of the hash function** – Determination of the hash function is always difficult in order for it to not influence the behaviour of the system and the data retrieval process. This problem is further exacerbated by the complex nature of the antigen, from which the hash function must be derived.
- **Potential for misclassification** - the hash function should not generate the same hash value for unrelated data items to avoid the issue increased false positives.
- **Collisions in the hash table** – reallocation in the hash table due to collisions will make identification of similar viruses even more difficult as the entries are spread throughout the table.
- **Limited capacity** – the size of the hash table will have to grow constantly to be able to handle the ever expanding list of viruses found, this may create problems for the hash function as it uses the table size as a parameter in its calculation of where to index and store the item.

However if the hash function can be created in such a way that viruses of a similar nature generate the same index reference then they will be stored in a linked list originating from the element in the data structure that the reference points to (using chaining). This would imply that each element in the initial hash table refers to a particular type of generic virus i.e. each element categorises a type of virus. Virus deviations from those specific types can then be identified by interrogating the linked list accessed from that element in the hash table.

In this way we can create a form of generalisation as we need only access the initial virus categories held in the hash table, and not the specific viruses in the linked lists off the hash table, to identify whether a presented item is a virus or not. This would be an ideal compromise for the IDS, however the hash function would likely be very complex to enable this functionality.

## 2.5 Case Based Reasoning. (CBR)

The CBR approach takes inspiration from the way in which human reasoning influences the decision making process. When faced with a problem an individual considers their past experiences and their previosuly used solutions and adapts those solutions given the circumstances of the new problem. In the context of CBR a case can be defined as "a contextualised piece of knowledge representing an experience"[21]. Each case normally comprises the following information [21]:
- The problem describing the state of the world when the case occurred
- The solution stating the derived solution to that problem and / or
- The outcome describing the state of the world after the case occurred

All these cases are then stored in a Case Base and used as a reference source in deciding upon a future action. The CBR process to generate new solutions from the case base occurs as follows [21]:
- Retrieve most similar cases to the current problem event.
- Reuse the case in an attempt to solve the current problem.
- Adapt the proposed solution if necessary.
- Store the newly identified solution as a new case in the case base.

Comprehensive work has been performed in identifying suitable ways to classify and represent an event or experience as a case. In addition complex tools have been developed to facilitate the efficient retrieval and comparison of relevant cases from the case base. As a result CBR would seem an ideal mechanism for memory within an IDS. Cases could be generated to reflect generic pathogenic sources. The complex retrieval algorithms currently available could then be used to interrogate the case base to find those cases that most closely match the newly introduced pathogen source. These cases could then be used as a starting point to investigate and classify the exact characteristics of the novel virus presented. Once classification has been successful the details of the novel virus can be incorporated in a new case and entered in the case base.

However a number of problems exist with this theory:

- **Issues regarding case representation:** In order to ensure meaningful comparisons are made between newly presented viruses and the case base of previously experienced viruses the attributes of the case have to be carefully considered to ensure all possible virus derivatives can be anticipated and accommodated. However envisioning the form of all future viruses and representing them in a series of specific attributes is an impossible task. Representation of cases is key in CBR systems and here that representation is made difficult due to the nature, diversity and evolution of the computer viruses we are trying to handle.
- **Retrieval and comparison difficulties**: Due to the difficulties raised in regard to representation, the retrieval process may select unsuitable cases for comparison; in addition cases that may be very similar but not have the same representation may be missed. This would lead to a misclassification of viruses in the system.
- **Human intervention**: The biggest issue with CBR systems, in terms of our application, is that the adaptation process relies on human intervention and is not automated or dynamic. Once the closest fitting case has been identified a number of adaptation mechanisms exist but most of these rely on human intervention to decide upon which adaptation is suitable and then introduce those changes. This is highly undesirable when compared to an AIS, where the system itself attempts to adapt (through mutation) the set of solutions it has to find a better fit.
- **Adaptation**: **Reliance on initial memory content**: The initial case base would need to incorporate a wide variety of possible virus classifications in order to facilitate the retrieval of meaningful cases for comparison. This implies the system has to have some form of initial memory manually created and entered in the system which in itself could influence the systems performance.
- **Updation of the memory set**: CBR uses the existing cases (at first those that are predetermined as the memory cases) to create a new case. Updation is therefore a manual process that relies heavily on those cases that exist in the current in the memory set. Others systems such as AIS's can start off with an empty memory set which is then dynamically created as the system experiences novel virus strains. The memory pool can therefore evolve in a more dynamic fashion than in CBR.

Considering the above issues it is clear that a CBR system in isolation is not suitable. The mechanism for storage and retrieval of information encapsulated in cases representing previously experienced events is attractive, however its only achievement would be to identify a set of previous viruses to which the system has had any experience. It would not provide any automated, dynamic means of adapting those virus profiles to more accurately reflect the currently experienced virus. In this way the learning and development properties of immune memory, desirable in such a system, would be limited by this mechanism.

A solution to this dilemma would be to develop a hybrid CBR system. A standard CBR system could be used to classify and retrieve generic virus types in the form of cases and then a dynamic unsupervised system such as an AIS could be used to autonomously mutate those cases in an attempt to improve the affinity of their fit to the current virus event. This system could generalise as only the generic virus classifications would be stored, reducing data redundancy and duplication. However the complex issues of virus representation within cases remain to be addressed.

## 2.6 Expert Systems

In order to solve a complex problem one would normally consult an expert in the field, as they have knowledge specific to that problem domain, leading to a quick and efficient resolution of the problem. However in most circumstances an expert my not be immediately available. To resolve this matter "expert systems are constructed by obtaining this knowledge from a human expert and coding it into a form that a computer may apply to similar problems". [22]

This knowledge is encapsulated in a number of facts and these facts are used to create a series of 'if – then' rules. The scenario can then be taken and examined in the context of the rules set created; a non expert would then be expected to reach the same diagnosis as the expert by identifying those rules that are satisfied by the scenario.

In the context of an IDS a computer virus expert could identify the facts associated with a particular electronic virus. These facts could be used to create a rule set that could diagnose such a virus from a presented scenario. Memory in this system would be encapsulated in the rules set as the rules reflect the facts associated with the specific viruses that are to be remembered. Each new virus would be examined against the rule set so they can be classified.

Unfortunately this system is inappropriate for our IDS application for the following reasons:

- **Inability to cope with novel viruses**: Viruses are constantly evolving in terms of their purpose, their form and construction. These novel virus candidates would be characterised by information which form novel facts that weren't in existence at the time the rule set was created. As no rules exist to accommodate these new facts it is highly likely that an expert system would fail to identify and classify such novel viruses as they lie outside the boundaries of the rules. Given that other systems (AIS's) are able to generalise and adapt to identify novel information this limitation seems fairly severe.
- **Reliance on an initial memory pool:** Unlike an AIS an Expert System is not able to dynamically develop a memory pool from an empty set, it needs to be told exactly what to remember at the point of its inception.

If-then rules provide a simple mechanism to classify potential computer viruses but due to the speed and variation in the way viruses are evolved such a rigid, inflexible system is wholly inappropriate.

## 3. Summary analysis.

We can summarise the above analysis in Table 1 by identifying the key attributes of memory and acknowledging which of the previously described implementations provide mechanisms to simulate the desirable properties that memory can offer, as discussed in Section 1.

A o symbol indicates the desirable property is observable from the system, a X indicates the property is not observable. ?'s reflect the situation where it is unknown as to whether the system exhibits this property.

|  | Complexity Theory | Neural Networks | AIS | | | Hash Tables | CBR | Expert Systems |
|---|---|---|---|---|---|---|---|---|
|  |  |  | Long Lived Cells | Idiotypic Network | Dynamic Population |  |  |  |
| Generalisation | o | o | o | o | o | X | o | X |
| Adaptation | o | o | o | o | o | X | X | X |
| Persistence | X | o | o | X | X | o | o | o |
| Scalability / capacity issues | X | X | X | X | X | X | o | X |
| Memory extraction | X | o | o | X | X | o | o | o |
| Unbiased | X | o | o/ X | o | ? | o | o | o |
| Accuracy | o/ X | o/ X | o/ X | o/ X | o/ X | o/ X | o/X | o/X |
| Ease of implementation | X | o | o | X | X | o | o | o |

**Table 1 Memory Analysis Matrix**

## 4. Conclusion

Hopefully it is clear from this analysis that memory represents an integral part of a system's functionality. It does not simply facilitate the storage and retrieval of information learnt by the system but provides a vital feedback mechanism that influences the learning process in that system. Simplifying the implementation of memory will therefore artificially constrain the potential functionality of that system.

Looking at the memory implementations discussed we can identify certain techniques that in isolation are clearly inappropriate for a memory mechanism in an IDS. CBR uses a logical approach and has desirable generalisation properties but requires an external source to drive the adaptation of its memory pool to recognise novel items upon presentation. In a similar manner, Expert Systems are too rigid and inflexible to offer a valuable alternative. They would require all viruses to be anticipated in advance in order to create rules to classify them; this is clearly impossible considering the evolutionary advancement of modern computer viruses. This inability to generalise or identify associations between novel items and the memory set would lead to a failure in the system to adequately classify items because the facts associated with the item presented would not conform to the system's rule set. Generalisation is also not possible with a standard Hash Table technique, as every single virus presented is simply stored in the system. Associations between common virus strains would not be made, leading to inefficiencies in the systems operation. If related viruses could be grouped together then common solutions could be applied on mass in a more efficient manner, but in the standard Hash Table methodology this is not possible.

Neural Networks have many advantages in terms of dynamic development, generalisation and association, being able to evolve to cope with novel information presented to them; however they take a long time to train and such systems have serious scalability issues. Given the scale of the IDS we are proposing such scalability problems may become a significant issue. In a similar vain Complexity Theory produces a dynamic, flexible memory solution that is able to generalise; however because memory is never specifically modelled, it emerges naturally as a property of the interactions of the entities within the system, then that memory may be extremely difficult to extract and use. It may generate the most versatile form of memory but if that memory cannot be extracted and analysed what value does it have.

This leaves us with immune inspired memory, as implemented in various AIS's. This form of memory is dynamic, it can generalise and is able to build associations between novel items presented and its existing memory pool. There are numerous alternative implementations in existence but at present they are simplistic in terms of memory utilisation and a desirable memory implementation would, by the nature of its natural counterpart, be complex. There are also issues with information being forgotten in some AIS's and also with the difficulty in being able to extract and utilise such memory, due to the way in which it is created and maintained in the system. However given all facts to consider this system offers the best and most flexible system for the IDS.

However a suitable solution may not be obtained by taking inspiration from a single system in isolation, some of the issues discussed could be resolved by creating a hybrid system. An example of such a hybrid system is to use the efficiency and generalisation properties of a CBR system whilst incorporating it with an AIS system whose purpose is to dynamically adapt the cases to autonomously evolve the case base so it can recognise novel virus strains. Alternatively the Hash Table function can be manipulated to create entries in the first hash table that represent generic virus strains. Those viruses that exist in linked lists that are referenced off those hash table entries then relate to the specific virus derivatives of those generic strains. In this way the hash table can be used to generalise.

Further work will be performed to investigate the potential for such hybrid systems as they represent the best opportunity to obtain the most desirable memory implementation that is suitable for an IDS.